\documentclass[pre,twocolumn,twoside,showpacs,superscriptaddress,floatfix]{revtex4-1}

\usepackage{graphics,amssymb,amsmath,epsf,color,hyperref}
\epsfclipon
\usepackage{epsfig,bm,epsf,graphics}
\usepackage{graphicx}
\usepackage{dcolumn}
\usepackage{bm}
\usepackage{braket}
\usepackage{color}
\usepackage{soul}
\usepackage{xcolor}
\usepackage{floatrow}
\usepackage{hyperref}

\usepackage[ruled,vlined]{algorithm2e}

\hypersetup{colorlinks, citecolor=blue, linkcolor=blue}

\usepackage{graphicx}

\begin{document}

\title{Stochastic Clock Attention for Aligning Continuous and Ordered Sequences}

\author{Hyungjoon Soh}
\email[Corresponding author: ]{hjsoh88@snu.ac.kr}
\affiliation{Department of Physics Education, Seoul National University, Seoul 08826, Korea}

\author{Junghyo Jo}
\email[Corresponding author: ]{jojunghyo@snu.ac.kr}
\affiliation{Department of Physics Education, Seoul National University, Seoul 08826, Korea}
\affiliation{Center for Theoretical Physics and Artificial Intelligence Institute, Seoul National University, Seoul 08826, Korea}
\affiliation{School of Computational Sciences, Korea Institute for Advanced Study, Seoul 02455, Korea}

\date{\today}

\begin{abstract}
We formulate an attention mechanism for continuous and ordered sequences that explicitly functions as an alignment model, which serves as the core of many sequence-to-sequence tasks. Standard scaled dot-product attention relies on positional encodings and masks but does not enforce continuity or monotonicity, which are crucial for frame-synchronous targets. We propose learned nonnegative \emph{clocks} to source and target and model attention as the meeting probability of these clocks; a path-integral derivation yields a closed-form, Gaussian-like scoring rule with an intrinsic bias toward causal, smooth, near-diagonal alignments, without external positional regularizers. The framework supports two complementary regimes: normalized clocks for parallel decoding when a global length is available, and unnormalized clocks for autoregressive decoding -- both nearly-parameter-free, drop-in replacements. In a Transformer text-to-speech testbed, this construction produces more stable alignments and improved robustness to global time-scaling while matching or improving accuracy over scaled dot-product baselines. We hypothesize applicability to other continuous targets, including video and temporal signal modeling.
\end{abstract}


\maketitle


\section{Introduction}
Sequence-to-sequence (Seq2Seq) learning addresses the fundamental problem of mapping a source sequence to a target sequence while preserving structure that emerges from the composition and ordering of atomic units (tokens, frames). Many real-world applications fit this paradigm, including neural machine translation~\cite{ref:NMT:bahdanau2015, ref:NMT:sutskever2014, ref:Transformer} and language modeling~\cite{ref:LM:bengio2003, ref:LM:brown2020, ref:LM:mikolov2010}, text-to-speech (TTS)~\cite{ref:Tacotron, ref:Tacotron2, ref:VITS, ref:AlignTTS, ref:DeepVoice3, ref:FastSpeech, ref:FastSpeech2, ref:Flowtron, ref:FlowTTS, ref:GlowTTS, ref:ParallelTacotron2, ref:Paranet, ref:TransformerTTS}, automatic speech recognition (ASR)~\cite{ref:ASR:wav2vec, ref:ASR:Whisper}, and music generation~\cite{ref:MusGen:hawthorne2019, ref:MusGen:jeong2019}. Classic solutions build on recurrent neural networks (RNNs)~\cite{ref:RNN} and gated variants such as LSTMs~\cite{ref:LSTM} and GRUs~\cite{ref:GRU}, which extend finite-order Markov dynamics via hidden-state recursion to capture longer-range temporal dependencies. In computer vision, analogous gains in long-range structure are achieved by stacking convolutional layers to expand the receptive field efficiently~\cite{ref:DConv}.

In temporal settings, long-range structure manifests as memory. RNN-based approaches broaden temporal capacity either by introducing explicit memory cells as in LSTM, where a cell state accumulates information and effectively introduces both short and long time scales~\cite{ref:LSTM}, or by composing multi-scale/pyramidal recurrent pathways that operate at different temporal resolutions and then aggregate features~\cite{ref:RNN:chung2016, ref:RNN:koutnik2014}. The introduction of attention mechanisms~\cite{ref:NMT:bahdanau2015} transformed this landscape by allowing models to form direct, learnable associations between distant elements, reducing reliance on a single recurrent channel for information transport. Transformer~\cite{ref:Transformer} architectures standardized this recipe -- multi-head self/cross-attention with position-wise feed-forward blocks -- and now constitute the default backbone for large-scale Seq2Seq systems across domains, including TTS~\cite{ref:TransformerTTS}. In causal generation, future masking prevents information flow from future to past, and positional encodings (absolute or relative) inject ordering and locality biases that vanilla dot-product attention lacks.

Despite these advances, many domains, especially frame-synchronous acoustics, impose stronger inductive biases than generic all-to-all similarity along several axes: (i) approximate monotonicity: the mapping from linguistic units to acoustic frames progresses forward in time, with large reorderings rare; violations manifest as skipped or repeated phones and unstable prosody; (ii) local continuity: adjacent acoustic frames are highly correlated, so abrupt attention jumps introduce audible artifacts (e.g., discontinuities, buzzing) and degrade the smoothness of predicted acoustic features; and (iii) strict causality: in autoregressive decoding, future information must not influence current predictions and should be enforced via causal masks or other hard constraints.

Prior work injects such structure through monotonic/online attention (e.g., MoChA~\cite{ref:Att:MoChA}), locality-aware~\cite{ref:Att:LocSensAtt} or position-aware attention~\cite{ref:Att:RelPosAtt}, and auxiliary losses that encourage near-diagonal encoder-decoder alignments~\cite{ref:Att:GuidPos}. While effective, these strategies can complicate optimization, be sensitive to hyperparameters, or still admit attention maps that intermittently violate continuity or monotonicity in practice. Moreover, in high-dimensional embedding spaces, unconstrained scaled dot-product attention can yield diffuse or unstable similarities unless guided by strong priors, so a diagonal (causal and continuous) structure is not guaranteed even with positional encodings.

We take a different route: we derive an attention scoring rule from a \emph{stochastic clock attention} (SCA) model evaluated with the Martin–Siggia–Rose / Onsager–Machlup path-integral formalism~\cite{ref:MSR, ref:MSR:Janssen, ref:MSR:DeDominicis, ref:OM, ref:OM2}. Clocks offer a principled way to encode the inductive biases of continuous, ordered targets: (i) a learned nonnegative accumulated rate induces a monotone, continuous time reparameterization, biasing attention toward causal, near-diagonal alignments without hand-crafted constraints; (ii) treating the clock as a stochastic process captures variability in local rate (e.g., speaking tempo) and yields data-dependent variance terms that modulate alignment confidence; and (iii) the path-integral treatment converts these ingredients into a meeting probability for two clocks and, under mild Gaussian assumptions, a closed-form Gaussian-like scoring rule with a Brownian-bridge variance profile. In this view, continuity, monotonic progression (``clocking''), and causality are built into an explicit alignment potential rather than appended as external regularizers. Concretely, we propose latent \emph{clocks} to source and target, evaluate their meeting kernel, and use the resulting logit—proportional to the negative squared clock difference normalized by the variance—to favor smooth, causal, near-diagonal alignments even without external positional embeddings.

This formulation complements, rather than replaces, modern architectures: we implement it as a drop-in cross-attention within a Transformer encoder-decoder~\cite{ref:Transformer} to isolate the effect of the attention rule. We evaluate two standard decoding regimes: (i) parallel decoding with normalized clocks, which assume a known or predicted global output length to map time to a unit interval and provide precise control over global pacing; and (ii) autoregressive decoding with unnormalized (diffusive) clocks, which operate causally and are trained with teacher forcing when length is unknown or generation is inherently stepwise. Considering both regimes enables a fair comparison to conventional scaled dot-product attention (SDPA) and broad applicability across training and inference scenarios.

To ground the theory in a realistic yet clean setting, we adopt a minimal TTS testbed where the targets are continuous mel-spectrograms and the source is a phoneme sequence with a strong ordering relation. Using the LJSpeech-1.1~\cite{ref:LJSpeech} dataset and a conventional front end, we compare our attention to the standard formulation under both parallel and autoregressive decoding regimes, holding all other architectural and optimization choices fixed. Our experiments focus on alignment behavior and downstream synthesis quality, providing evidence that principled inductive structure encoded directly in the attention potential can improve stability and faithfulness without relying on brittle heuristics or heavy regularization.

\section{Theory}

We begin by casting alignment attention as a path-integral probability kernel over stochastic clock trajectories. Specifically, we attach latent clocks $\lambda^X_s$ and $\lambda^Y_t$ to the source and target, and define a meeting kernel that gives the probability the two clocks coincide at $(s,t)$. Row-normalizing the resulting log-likelihood yields attention weights; equivalently, the attention logits are the log meeting probability up to a row-wise (i.e., $s$-dependent) additive constant. We first derive this construction for normalized clocks $\lambda\!\in\![0,1]$, which couple local time to the global endpoint and naturally support parallel decoding. Under a small-fluctuation linearization of the clock functionals, the score admits a closed form: a Gaussian potential in the clock difference with a Brownian-bridge–shaped variance. We then obtain the same score without normalization, yielding unnormalized clocks whose variance grows with length and that are compatible with autoregressive decoding.

Let \(X_s\) for \(s\!\in[0,S]\) and \(Y_t\) for \(t\!\in[0,T]\) denote continuous-time sequences, possibly with different dimensionalities or lengths. We formulate alignment as an expectation over feasible monotonic paths \(t(\cdot)\):
\[
L \;=\; \mathbb{E}_{\Pi}\!\big[L[X_s,\,Y_{t(s)}]\big]
\;=\; \int_{\mathcal{P}} L\!\big[X_s,\,Y_{t(s)}\big]\, d\Pi\!\big[t(\cdot)\big],
\]
where the path space is \(\mathcal{P}=\{t:[0,S]\!\to\![0,T],\  t(0)=0,\ t(S)=T\}\) with each $t(\cdot)$ absolutely continuous and non-decreasing. The path measure $\Pi$ induces a pointwise alignment intensity, given by
\[
P(s,t)\ \propto\ \int_{\mathcal{P}} \delta\!\big(t - t(s)\big)\, d\Pi\!\big[t(\cdot)\big],
\]
which represents the marginal probability of aligning position $s$ in $X$ with time $t$ in $Y$.
In a neural model, this induces attention weights: for each query $s$, the attention over $t$ is modeled as \(\mathrm{Attn}(s,\cdot)=\mathrm{softmax}_t\big(\mathrm{Score}(s,t)\big)\) with $\mathrm{Score}(s,t)$ chosen proportional to $P(s,t)$.

Assume \(X\) and \(Y\) share a latent internal \emph{clock} that governs their alignment. Let \(\eta^X_s=\mathcal{F}(X_s)\) and \(\eta^Y_t=\mathcal{G}(Y_t)\) be learned feature projections. We define \emph{normalized clocks} by accumulating a nonnegative function \(\phi(\cdot)\):
\begin{equation}
\label{eq:clock}
\lambda^X_s \;=\; \frac{\int_0^s \phi(\eta^X_u)\,du}{\int_0^S \phi(\eta^X_u)\,du},
\qquad
\lambda^Y_t \;=\; \frac{\int_0^t \phi(\eta^Y_v)\,dv}{\int_0^T \phi(\eta^Y_v)\,dv},
\end{equation}
so that $\lambda^X_s$ and $\lambda^Y_t \in[0,1]$ are continuous and ordered time reparameterizations of time. To ensure strict monotonicity, we require \(\phi(\cdot)\ge \varepsilon>0\), for example by adding a small constant \(\varepsilon\). Typical choices for $\phi(x)$ include \(e^{x}\) or the smooth Softplus function \(\log(1+e^{\beta x})/\beta\).

If two normalized clocks coincide, \(\lambda^X_s=\lambda^Y_t\), then the pair \((s,t)\) lies on a candidate alignment path. To formalize this, we introduce the \emph{meeting kernel}
\begin{equation}
\label{eq:kernel}
K_{\mathrm{meet}}(s,t)\;=\;\mathbb{E}\!\left[\delta\!\big(\lambda^X_s-\lambda^Y_t\big)\right],
\end{equation}
which measures the probability density of the clocks meeting at $(s,t)$. We evaluate this kernel using the Martin–Siggia–Rose / Onsager–Machlup formalism~\cite{ref:MSR, ref:OM}, a path-integral framework well suited to stochastic clocks because each clock is a time integral of a noisy rate (the learned projection passed through $\phi$), so its trajectory is inherently random. Modeling these rate fields as (approximately) Gaussian processes makes the MSR/OM action quadratic after a first-order (small-fluctuation) linearization, turning the path probability into a tractable Gaussian functional integral.

Specifically, we decompose the learned projections into mean and fluctuation terms,
\[
\eta^X_s=\eta^x_s+\xi^X_s,\qquad \eta^Y_t=\eta^y_t+\xi^Y_t,
\]
where $\xi^X$ and $\xi^Y$ are independent zero-mean Gaussian fields with covariances \(C_X(u,v)=\mathbb{E}[\xi^X_u\xi^X_v]\), \(C_Y(u,v)=\mathbb{E}[\xi^Y_u\xi^Y_v]\). For simplicity, we focus on the independent case and ignore any cross-correlation terms.

In this formalism, fluctuations are weighted by quadratic {\it actions}. For $X$, the Onsager-Machlup weight takes the form
\begin{align*}
P[\xi^X]&\propto \exp\Big({-S_X[\xi^X]}\Big),\\
S_X[\xi^X]&=\tfrac12\!\int_0^S\!\!\int_0^S \xi^X_u\,C_X^{-1}(u,v)\,\xi^X_v\,du\,dv,
\end{align*}
where $S_X$ serves as the {\it action} in the path-integral formulation, penalizing deviations according to the inverse covariance operator $C_X^{-1}$. An analogous expression holds for \(Y\). We assume throughout that $C_X$ and $C_Y$ are invertible on the relevant subspaces.

The normalized clocks are smooth functionals of the projected sequences. For small noise amplitude, we obtain the variational path by linearizing around the deterministic trajectories $\eta^x$ and $\eta^y$:
\begin{align*}
\delta \lambda^X_s& =\int_0^S J^x_s(u)\,\xi^X_u\,du,\\
J^x_s(u)&=\frac{\big(\mathbf{1}_{\{u\le s\}}-\lambda^x_s\big)\,\phi'(\eta^x_u)}{\int_0^S \phi(\eta^x_w)\,dw},
\end{align*}
with an analogous expression for \(\delta\lambda^Y_t\) using \(J^y_t\).

Defining the deterministic offset \(\Delta_{s,t}=\lambda^x_s-\lambda^y_t\) and the perturbed difference \(Z_{s,t}=\Delta_{s,t}+(\delta\lambda^X_s-\delta\lambda^Y_t)\), we evaluate the meeting kernel of Eq.~(\ref{eq:kernel}) using the Fourier representation of the delta function, \(\delta(z)=\int \tfrac{dk}{2\pi}\exp(ikz)\), together with  the Gaussian moment-generating functional:
\begin{align*}
 \mathbb{E}\!\left[\exp\!\left( ik\!\Big(\int J^x_s\xi^X-\int J^y_t\xi^Y\Big) \! \right)\right] \\
=\exp\!\Big(-\tfrac12 k^2\,[A_X(s)+A_Y(t)]\Big),
\end{align*}
where the variance contributions are
\begin{align*}
A_X(s)&=\int_0^S\!\!\int_0^S J^x_s(u)\,C_X(u,v)\,J^x_s(v)\,du\,dv,\\
A_Y(t)&=\int_0^T\!\!\int_0^T J^y_t(u)\,C_Y(u,v)\,J^y_t(v)\,du\,dv.
\end{align*}
Using the moment-generating functional, integrating over $k$ yields a Gaussian form for the meeting kernel:
\begin{align*}
K_{\mathrm{meet}}(s,t\,|\,\eta^x,\eta^y)
=\frac{1}{\sqrt{2\pi\,\Sigma_{s,t}^2}}\,
\exp\!\left(-\frac{\Delta_{s,t}^2}{2\,\Sigma_{s,t}^2}\right),
\end{align*}
where $\Sigma_{s,t}^2=A_X(s)+A_Y(t)$.
Since attention weights are normalized through a softmax in \(t\), we adopt the log-likelihood associated with Eq.~(\ref{eq:kernel}) as the scoring function. This gives the score of stochastic clock attention,
\begin{equation}
\label{eq:score}
\mathrm{Score}(s,t)=
-\frac{\big(\lambda^x_s-\lambda^y_t\big)^2}{2\,\Sigma_{s,t}^2}+C,
\end{equation}
where we omit the \(-\tfrac12\log \Sigma_{s,t}^2\) term because it is approximately row-wise constant and gets absorbed by the softmax normalization.

To make $\Sigma_{s,t}^2$ explicit, we approximate $\phi(\eta^X)$ as a stationary process with mean $\mu_X=\mathbb{E}[\phi(\eta^X)]$ and integrated covariance
$$
K_X \;=\; \int_{-\infty}^{\infty}\!\mathrm{Cov}\!\big(\phi(\eta^X_0),\phi(\eta^X_\tau)\big)\,d\tau,
$$
and analogous definitions hold for $\mu_Y$ and $K_Y$. Using a mean–variance surrogate approximation, the variance contribution becomes
\begin{equation}
A_X(s)\;\approx\;\frac{1}{S}\frac{K_X}{\mu_X^{2}}\frac{s}{S}\!\left(1-\frac{s}{S}\right),
\label{eq:variance}
\end{equation}
with a similar expression for $A_Y(t)$.
The quadratic profile in Eq.~(\ref{eq:variance}) is characteristic of Brownian bridge, which vanishes at the endpoints and attains maximal variance near the midpoint.

This arises from a first-order delta-method linearization. Equivalently, by treating the integrated fluctuations as Gaussian and applying Itô calculus to the ratio,
$$
\lambda^X_s \;\approx\; \frac{\mu_X s+\sqrt{K_X}\,W^X_s}{\mu_X S+\sqrt{K_X}\,W^X_S}
\;\approx\; \frac{s}{S} \;+\; \frac{\sqrt{K_X}}{\mu_X S}\,B^X_s,
$$
where $W^X_s=\int_0^s \!\big(\phi(\eta^X_u)-\mu_X\big)\,du$ is an integrated fluctuation process, and $B^X_s=W^X_s-\tfrac{s}{S}W^X_S$ is a Brownian bridge constrained to vanish at both endpoints. If $\eta^X$ and $\eta^Y$ share the same statistics and the same nonlinearity $\phi$, then the shape parameters $\mu_*$ and $K_*$ coincide and can be absorbed into a constant. This approximation allows $\Sigma^2_{s,t}$ to be evaluated in closed form, without computing the full nonlinear variance term.

Normalized clocks couple local time to the endpoint through their denominator, meaning that computing $\lambda$ requires access to the full sequence (i.e., global lengths $S$ and $T$). This construction naturally aligns with \emph{parallel} decoding. By contrast, autoregressive decoding requires clocks that depend only on local information. The simplest approach is to remove the normalization, yielding the {\it unnormalized clocks}
$$
\tilde{\lambda}^X_s \;=\; \int_0^s \phi(\eta^X_u)\,du,
\qquad
\tilde{\lambda}^Y_t \;=\; \int_0^t \phi(\eta^Y_v)\,dv.
$$
To compare them on a dimensionless scale, we rescale via $\bar{\lambda}^X_s:=\tilde{\lambda}^X_s/(\mu_X S)$.
Under the same approximation as before, the variance becomes
\begin{equation}
    \tilde{A}_X(s)=\mathrm{Var}(\bar{\lambda}^X_s)\approx\frac{1}{S}\frac{K_X}{\mu_X^{2}}\frac{s}{S}
\label{eq:variance-unnorm}
\end{equation}
with an analogous expression for $Y$.
Thus, the score in Eq.~\eqref{eq:score} retains the same functional form, but the effective variance $\Sigma_{s,t}^2$ now scales linearly with sequence length. This variant is naturally compatible with causal and step-wise (autoregressive) attention updates that do not require global length information.

By making three simplifying assumptions, our unnormalized clock attention reduces to conventional scaled dot-product attention: (i) eliminate the clock constraint by identifying the temporal parameter with the raw fluctuation, $\lambda_s^X=\eta_s^X$; (ii) treat the variance as time-independent, $\Sigma_{s,t}^2=1$; and (iii) replace the squared $L_2$ distance with the dot-product similarity and absorb the (approximately constant) norms into the scale. The resulting attention score is
$$
\mathrm{SDPA\_Score}(s,t)\;=\;\frac{(\eta_s^X)^{\top}\eta_t^Y}{\sqrt{d}},
$$
i.e., the usual scaled dot-product with the $\sqrt{d}$ dimensional scaling.

\section{Experiments}
\subsection{Algorithm}

\begin{algorithm}[ht]
\caption{Stochastic Clock Attention (SCA)}
\label{alg:clock-attn}
\SetKwInOut{KwIn}{Input}\SetKwInOut{KwOut}{Output}
\KwIn{
  $q\!\in\!\mathbb{R}^{B\times L_q\times D}$, $k\!\in\!\mathbb{R}^{B\times T_k\times D_k}$, $v\!\in\!\mathbb{R}^{B\times T_k\times D_v}$,
  $q\_kpm\!\in\!\{0,1\}^{B\times L_q}$, $k\_kpm\!\in\!\{0,1\}^{B\times T_k}$,
  $\epsilon>0$, \texttt{normalize} $\in\{\texttt{true},\texttt{false}\}$
}
\KwOut{$o\!\in\!\mathbb{R}^{B\times L_q\times D}$, $A\!\in\!\mathbb{R}^{B\times L_q\times T_k}$}
\SetKwFunction{MTN}{MaskedTimeNorm}
\SetKwFunction{Clock}{Clock}
\SetKwFunction{CDS}{ClockDiffScore}
\SetKwProg{Fn}{Function}{:}{end}

\Fn{\MTN{$x,\; mask$}}{
  $count \gets \sum mask$\;
  $\mu \gets \dfrac{\sum (x\odot mask)}{count}$;\quad
  $v \gets \dfrac{\sum ((x-\mu)^2\odot mask)}{count}$\;
  $z \gets \dfrac{x-\mu}{\sqrt{v+\epsilon}}$\;
  \KwRet $z\odot mask$\;
}

\Fn{\Clock{$x,\; mask,\; \texttt{normalize}$}}{
  $pm \gets mask_{:,0{:}L-1}\land mask_{:,1{:}L}$\;
  $y \gets \tfrac12\big(x_{:,0{:}L-1}+x_{:,1{:}L}\big)$\;
  $g \gets (\phi(y)+\epsilon)\odot pm$\;
  $z_0 \gets \text{pad}(\text{cumsum}(g),0)$\;
  \uIf{\texttt{normalize}}{
    $z \gets z_0 / \sum g$\;
    $pos \gets \big(\text{cumsum}(mask)-0.5\big)/\sum mask$\;
    $var \gets pos(1-pos)$\;
  }\Else{
    $z \gets z_0$\;
    $pos \gets \text{cumsum}(mask)-0.5$\;
    $var \gets pos$\;
  }
  \KwRet $(z\odot mask,\; var)$\;
}

\Fn{\CDS{$\eta^X_s,\eta^Y_t,s\_mask,t\_mask,\epsilon,\texttt{normalize}$}}{
  $len\_s \gets \sum s\_mask$;\quad $len\_t \gets \sum t\_mask$\;
  $(\lambda_s,\; var_s) \gets \Clock(\eta_s^X,\; s\_mask,\; \texttt{normalize})$\;
  $(\lambda_t,\; var_t) \gets \Clock(\eta_t^Y,\; t\_mask,\; \texttt{normalize})$\;
  $\Sigma^2 \gets var_s/len\_s \;+\; var_t^\top/len\_t$\;
  $A \gets \sum \lambda_s^2$;\quad $B \gets (\sum \lambda_t^2)^\top$;\quad $D \gets \lambda_s \lambda_t^\top$\;
  $dist^2 \gets A + B - 2D$\;
  $S \gets -\,dist^2\,/\,(2\sqrt{d}\,\Sigma^2+\epsilon)$\;
  \KwRet $S$\;
}

$allow \gets \text{expand}(q\_kpm,2)\land \text{expand}(k\_kpm,1)$\;
$\eta_s^X \gets \MTN(qW_q,\; q\_kpm)$\;
$\eta_t^Y \gets \MTN(kW_k,\; k\_kpm)$\;
$V \gets vW_v$\;
$Score \gets \CDS(\eta_s^X, \eta_t^Y, q\_kpm, k\_kpm, \epsilon, \texttt{normalize})$\;
$logits \gets Score \cdot \text{logit\_scale}$\;
$logits[\lnot allow] \gets -\infty$\;
$A \gets \text{softmax}(logits)$;\quad $o \gets A\,V$\;
\KwRet $(o, A)$\;
\end{algorithm}

We implement stochastic clock attention (SCA) as an attention module paired with a scoring function \textsc{ClockDiffScore} (Eq.~\eqref{eq:score}) that returns a negative energy used directly as the attention logit. Queries and keys are first standardized along time with \textsc{MaskedTimeNorm} so per-timestep means/variances remain well-behaved under padding, improving gradient stability and preventing padded spans from biasing statistics. As summarized in Algorithm~\ref{alg:clock-attn}, padded-valid masks from queries and keys are combined into an ``allow'' mask (with causal mask included in AR training); projected features are normalized $\eta_s^X=\textsc{MaskedTimeNorm}(qW_q)$, $\eta_t^Y=\textsc{MaskedTimeNorm}(kW_k)$; and values are linearly mapped $V=vW_v$. Clocks are constructed for both sequences by averaging features on mid-edges (with corresponding masks), applying a nonnegative function $\phi$ with a small $\epsilon$ to ensure strict positivity, cumulatively summing with a left-zero pad (anchoring at $0$), and scaling to $[0,1]$ when using normalized clocks (Eq.~\eqref{eq:clock}). We use $\phi(x)=\tfrac12\!\left(1+x\frac{1+x+|x|}{1+|x|}\right)$, which resembles Softplus but is less prone to gradient vanishing. This yields clock trajectories $\lambda_s$ and $\lambda_t$ and a variance surrogate $var$ as a function of the normalized position $pos$: for normalized clocks, $var\propto pos(1{-}pos)$ (Brownian-bridge profile; Eq.~\eqref{eq:variance}); for unnormalized clocks, $var\propto pos$ (diffusive profile; Eq.~\eqref{eq:variance-unnorm}).

The denominator of the energy uses a length-aware combination $\Sigma^2 = \frac{var_s}{len_s} + \frac{var_t^\top}{len_t}$, while the numerator uses the squared clock difference computed efficiently via norms and a Gram product to avoid materializing a $B\times S\times T\times d$ tensor: $\|\lambda_s-\lambda_t\|_2^2 = \|\lambda_s\|_2^2 + \|\lambda_t\|_2^2 - 2\,\lambda_s^\top\lambda_t$. The resulting score is
$$
S \;=\; -\,\frac{\|\lambda_s-\lambda_t\|_2^2}{\,2\sqrt{d}\,\Sigma^2}\,,
$$
which behaves like a Gaussian potential in the clock difference and is finally scaled by \texttt{logit\_scale} to control sharpness; invalid pairs are masked to $-\infty$, a row-wise softmax over keys produces $A$, and the context is $o=AV$. Normalized score implements the parallel-decoding variant that respects a global target length, whereas setting \texttt{normalize} to \texttt{false} yields a unnormalized clock suitable for autoregressive decoding. Dividing by $\sqrt{d}$ stabilizes the initial magnitude of scores across feature dimensions, and constants shared between source and target variance are absorbed into \texttt{logit\_scale}.

\subsection{TTS Experiments}

We evaluate stochastic clock attention formalism on a simple text-to-speech (TTS) task, since TTS maps discrete linguistic tokens to continuous acoustic features under a strong, approximately monotonic temporal correspondence between source and target.

In sequence-to-sequence TTS, two training/decoding regimes dominate: parallel (non-AR) and autoregressive (AR). Parallel decoders synthesize multiple (often all) frames simultaneously and usually rely on explicit or implicit alignment/length mechanisms; examples include FastSpeech~1/2~\cite{ref:FastSpeech, ref:FastSpeech2}, Flow-TTS~\cite{ref:FlowTTS}, Glow-TTS~\cite{ref:GlowTTS}, and AlignTTS~\cite{ref:AlignTTS}. In contrast, AR models generate one frame at a time, typically with teacher forcing during training, so attention must be computed frame-wise. Representative AR systems include Tacotron~1/2~\cite{ref:Tacotron, ref:Tacotron2}, Deep Voice 3~\cite{ref:DeepVoice3}, TransformerTTS~\cite{ref:TransformerTTS}, and Flowtron~\cite{ref:Flowtron}.

Our stochastic clock attention admits two instantiations. With normalized clocks, the clock functions encode global timing over the entire utterance; clocks must be computed for the full time range {\it a priori}, which therefore restricts decoding to the parallel setting. With unnormalized clocks, the clocks do not carry explicit length information; this allows autoregressive decoding while preserving the core principle that a clock is the integral of an underlying random field.

We compare our attention to the conventional scaled dot-product multi-head attention under both decoding regimes: (i) parallel decoding with normalized clocks, where we predefine the global length to instantiate decoder queries; and (ii) an autoregressive baseline trained with teacher forcing, where the decoder consumes the previous mel frame and predicts the next with a causal mask. This design isolates the contribution of the attention mechanism from that of the decoding strategy.

We train and evaluate on the LJSpeech-1.1~\cite{ref:LJSpeech} corpus, a single-speaker English audiobook dataset comprising 13,100 recordings ($\approx$ 24 hours) sampled at 22.05 kHz. For acoustics, each waveform is converted to 80-dimensional mel-spectrograms using PyTorch’s \texttt{torchaudio MelSpectrogram} with parameters $n_{\rm mels}=80$, $n_{\rm FFT}=1024, hop_len=256, win_len=1024$ and follow the library defaults to ensure reproducibility. On the linguistic side, input text is first punctuation-normalized and then mapped from graphemes to phoneme sequences using a standard Python English Grapheme-to-Phoneme (G2P) module; the resulting phoneme tokens serve as the sole linguistic features fed to the model. This setup -- (i) a widely used single-speaker dataset, (ii) a conventional 80-bin mel front end with off-the-shelf parameters, and (iii) phonemic inputs derived via automated G2P -- provides a clean, replicable baseline for analyzing model behavior independent of dataset idiosyncrasies or handcrafted alignments.

Our testbed is a Transformer~\cite{ref:Transformer} encoder-decoder that consumes phonemes as linguistic inputs and predicts mel-spectrograms as acoustic outputs. The encoder is a 6-layer Transformer with PyTorch’s default multi-head attention; the decoder is a 4-layer Transformer whose self-attention likewise uses the default multi-head module ($d_{\rm model}=256,\  n_{\rm head}=4,\ dropout=0.1$ with sinusoidal positional encoding). We keep all architectural and optimization hyperparameters fixed across conditions and change only the cross-attention operator to evaluate our proposed mechanism. Throughout, the encoder processes the phoneme sequence, while the decoder operates in mel space.

\begin{figure*}[t]
\includegraphics[width=\columnwidth]{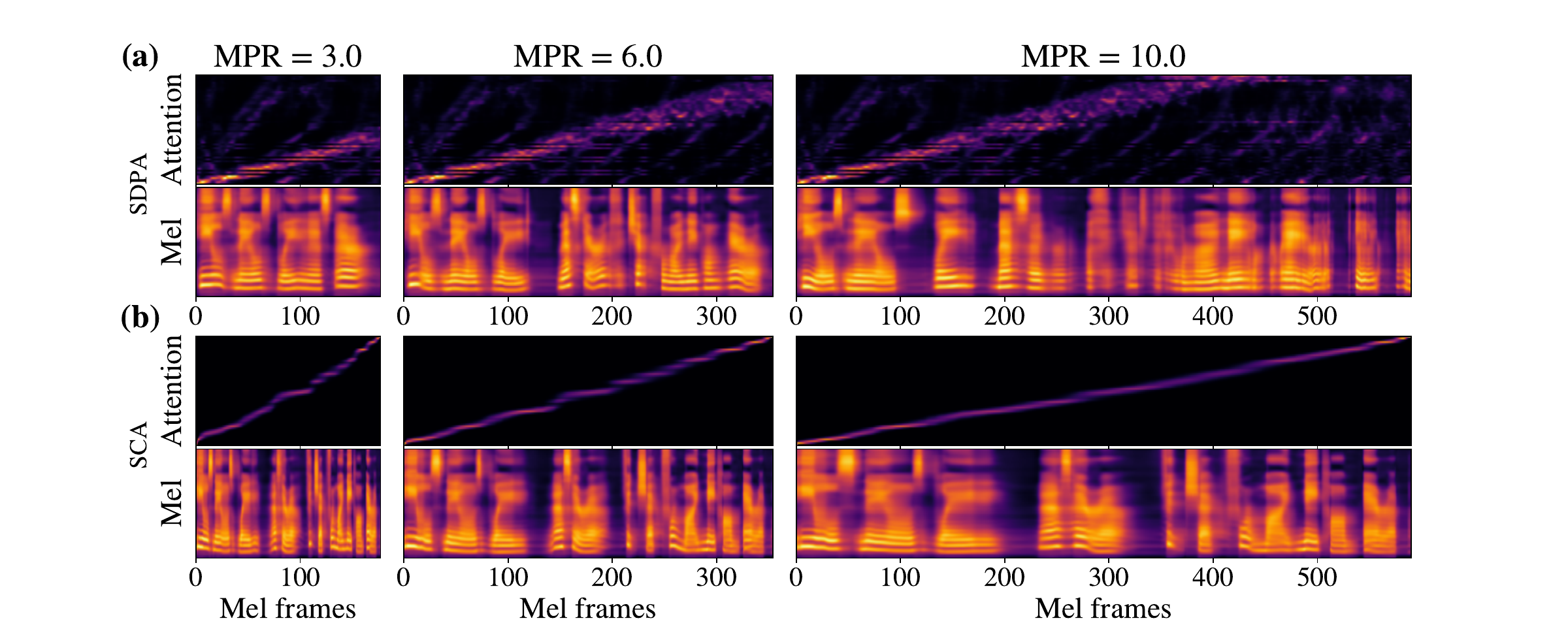}
    \centering
    \caption{
        Comparison of attention maps and generated mel-spectrograms from sample text, ``{\it Heels, nails, blade, mascara, fit check for my napalm era.}'', for parallel-decoding models with two cross-attention mechanisms: scaled dot-product (a) and normalized clock (b). For a fixed mel-frame scale, decoding length is controlled by the mel-to-phoneme ratio (MPR), shown for MPRs of 3.0 (top), 6.0 (middle), and 10.0 (bottom).
    }
\label{fig:ParallelEx}
\end{figure*}

\begin{figure}[h]
\includegraphics[width=0.65\columnwidth]{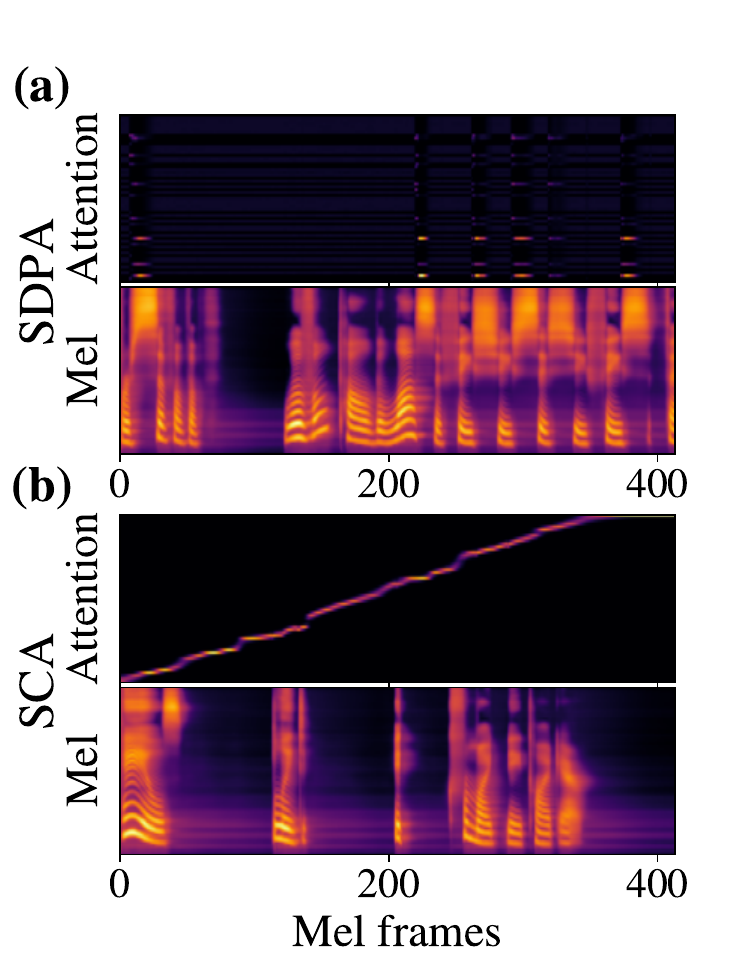}
    \centering
    \caption{
        Comparison of attention maps and generated mel-spectrograms for the sample text ``{\it Heels, nails, blade, mascara, fit check for my napalm era.}'' produced by autoregressive decoding models with two cross-attention mechanisms: scaled dot-product (a) and unnormalized clock (b). The maximum mel-frame length is constrained by the mel-to-phoneme ratio (MPR) set to 7.0.
    }
\label{fig:AREx}
\end{figure}

\begin{figure}[h]
\includegraphics[width=\columnwidth]{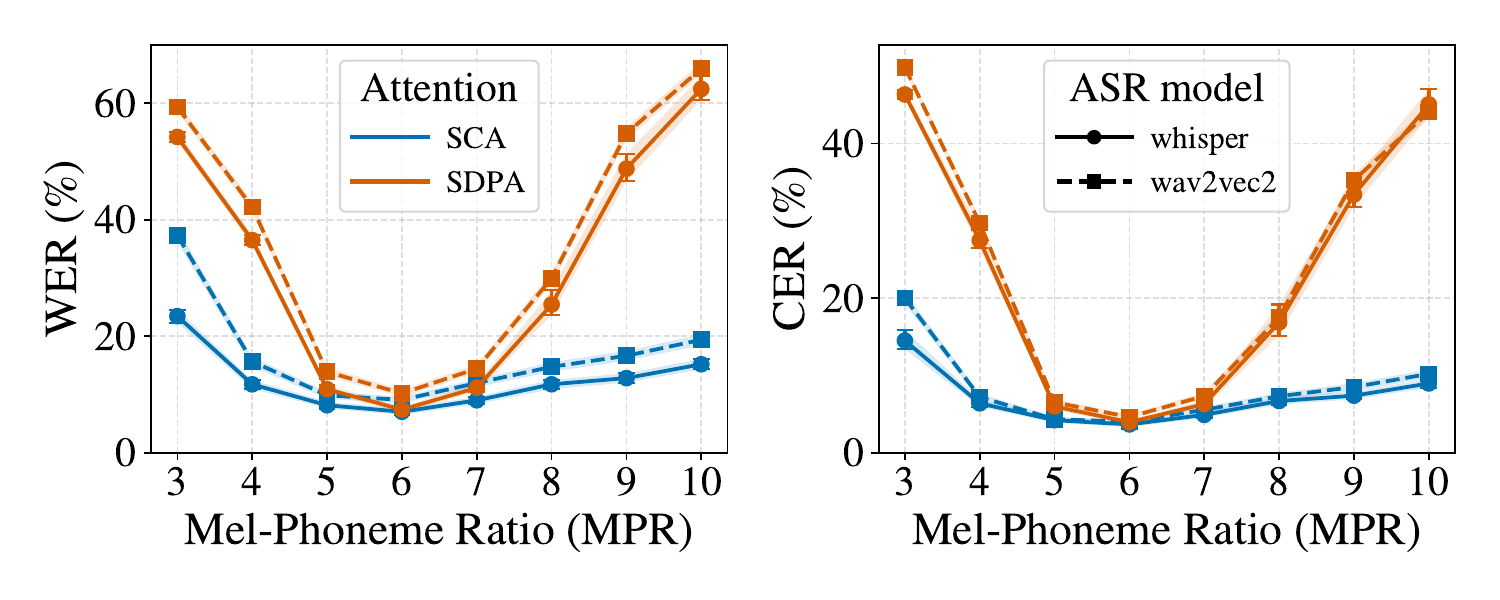}
    \centering
    \caption{
        Performance of TTS models using scaled dot-product attention (SDPA, orange) vs. stochastic clock attention (SCA, blue), evaluated with two ASR back-ends: Whisper (circles) and wav2vec 2.0 (squares). Left: word error rate (WER). Right: character error rate (CER). Points show means; error bars denote 95\% confidence intervals obtained from 1,000 bootstrap resamples of sentences (lower is better).
    }
\label{fig:Scores}
\end{figure}

We compare four configurations that share architecture and optimization settings except for the attention score and decoding regime: normalized-clock attention with parallel decoding, scaled dot-product attention with parallel decoding, unnormalized clock attention with autoregressive decoding, and scaled dot-product attention with autoregressive decoding. All models are trained on a single NVIDIA RTX 3090 (24 GB) using AdamW optimizer~\cite{ref:AdamW} with a learning rate $10^{-4}$ and batch size 48 and $L_1$ distance on mel-spectrograms as loss; all models have 8.72M parameters and are trained for 2,000 epochs.

We evaluate on the union of the CMU ARCTIC prompt set~\cite{ref:ARCTIC, ref:ARCTIC2} and Harvard Sentences~\cite{ref:HarvardSentences} (1,852 sentences total). Speech is synthesized with our TTS models and an external HiFi-GAN vocoder~\cite{ref:HifiGAN} trained with the same acoustic configuration and dataset as used during TTS training; the vocoder is fixed across all experiments. Recognition is performed by two ASR back-ends: Whisper~\cite{ref:ASR:Whisper} and wav2vec 2.0 with CTC~\cite{ref:ASR:wav2vec, ref:CTC}, and we report word error rate (WER) and character error rate (CER). Unless otherwise noted, we use each ASR system’s default decoding settings.

For parallel (non-autoregressive) models, we sweep the target mel-to-phoneme ratio (MPR = target mel frames / input phoneme count) over $\{3.0,4.0,\dots,10.0\}$ to probe robustness to global time-scaling; the training-set mean MPR is $\approx 6.6$. For the autoregressive model, the maximum mel-frame length is constrained by the mel-to-phoneme ratio (MPR) set to 7.0.

We show qualitative behavior under parallel decoding in Fig.~\ref{fig:ParallelEx}. We visualize attention maps and mel outputs for the sample text ``{\it Heels, nails, blade, mascara, fit check for my napalm era.}'', at target lengths $\mathrm{MPR}\in\{3.0,6.0,10.0\}$. The scaled dot-product baseline exhibits an almost constant average pace across target lengths: at short ratios, synthesis truncates before covering all phonemes; at long ratios, generation over-extends with increasing attention blur and degraded ASR scores. In contrast, normalized-clock attention adapts its pace to the specified global length while maintaining sharper, near-diagonal alignments, yielding recognizable speech across ratios with relatively stable WER/CER.

Figure~\ref{fig:AREx} shows corresponding results for the autoregressive decoder. With scaled dot-product attention, the model often fails under our setup to form coherent temporal alignments and does not synthesize intelligible audio: despite high-fidelity acoustics, the phonemes are essentially random, and both WER and CER approach 100\%. Under identical settings, unnormalized clock attention produces diagonal and continuous alignments, and yields partially intelligible speech: the pauses and speaking rate are unstable, but the phonemes are largely in the correct order.

Finally, aggregated accuracy is shown in Fig.~\ref{fig:Scores}. Over the 1,852-sentence set, normalized-clock attention under parallel decoding matches or exceeds the scaled dot-product baseline and shows lower variability as the global length changes. At the best shared MPR ($6.0$), normalized-clock attention achieves $7.03\%$ WER and $3.66\%$ CER, compared with $7.39\%$ WER and $3.94\%$ CER for scaled dot-product attention. In the autoregressive regime, the scaled dot-product model often fails across sentences under our setup, whereas unnormalized clock attention yields intelligible speech on a majority of prompts with $66.5\%$ WER and $48.5\%$ CER on the same evaluation set. In this testbed, parallel decoding outperforms autoregressive decoding, plausibly because global length information is provided during both training and inference. Notably, normalized-clock attention maintains a broadly recognizable regime (WER $<20\%$) across a wide range of generated MPRs, demonstrating effective speed control.

To summarize, parallel decoding with normalized clocks is effectively length-equivariant: intelligibility and attention sharpness remain stable as target length varies. We also observe a mid-trajectory softening of attention-consistent with a Brownian-bridge-like variance term: which may encourage exploration and prevent overly rigid diagonal paths. Quantifying this generalization-confidence trade-off is left for future work.


\section{Conclusion}

We introduced an attention-as-alignment formulation for pairs of sequences that exhibit a strict ordering relation. The key idea is to encode monotonic progression through random clocks: learned and nonnegative accumulators whose (normalized) integrals reparameterize time on the source and target. Using a path-integral view, we derived a closed-form meeting kernel over clock differences, yielding an analytic attention score that favors near-diagonal, continuous, and causal alignments. The same derivation naturally supports two decoding regimes: a parallel variant using normalized clocks (requiring a global length to instantiate queries), and an autoregressive variant using unnormalized clocks (no global length needed).

We validated the approach in a minimal TTS testbed, holding all architecture and optimization choices fixed except for the cross-attention rule. In parallel decoding, the proposed score produced attention maps that were more stable over long utterances and led to robust generation when the total frame count (and thus speaking rate) was varied -- i.e., consistent content under global time-stretch. In the autoregressive setting, the same principle applied without normalization yielded more temporally consistent synthesis when decoded frame-by-frame. Objective checks via a downstream ASR system supported these trends.

A practical advantage is that our method adds no trainable parameters and requires only a change to the attention logit computation. Its inductive bias is most suitable when targets are approximately time-warp invariant (e.g., mel-spectrograms, where modest changes in length preserve content up to resampling). By contrast, for discrete token generation where changing the length can alter semantics, an external length model (or a compatible global-length predictor) is needed to exploit the parallel variant. Two limitations are worth noting: the approach presumes monotonic alignments and, in its normalized form, benefits from access to (or reliable prediction of) the global output length.

Looking ahead, we plan to extend the clocks to multi-scale and hierarchical versions so that variability in local tempo and long-range pacing can be modeled jointly combining coarse ``global clocks'' with fine ``local clocks'' to model heterogeneous speed and variance across long sequences. We also aim to connect the alignment potential to diffusion/flow matching decoders, where a clock-informed guidance term could stabilize synthesis without heavy regularizers. Finally, because the formulation directly rewards smooth and causal alignments given a known or controllable length, we see immediate opportunities in video~\cite{ref:Sora} and musical performance generation~\cite{ref:MusGen:hawthorne2019, ref:MusGen:jeong2019}, where reducing attention discontinuities can mitigate visual or auditory ``jumps'' during long sequences.

\section*{Acknowledgment}
This work was supported by the Creative-Pioneering Researchers Program through Seoul National University, and the National Research Foundation of Korea (NRF) grant (Grant No. 2022R1A2C1006871) (J.J.)

\bibliographystyle{apsrev4-1}
\bibliography{reference}

\end{document}